\pdfoutput=1

\documentclass[11pt]{article}

\usepackage[final]{ACL2023}

\usepackage{times}
\usepackage{latexsym}

\usepackage[T1]{fontenc}

\usepackage[utf8]{inputenc}

\usepackage{microtype}

\usepackage{inconsolata}

\usepackage{graphicx}
\usepackage[ruled]{algorithm2e}
\usepackage{amsfonts, amssymb}
\usepackage{multirow}

\title{Label-Free Multi-Domain Machine Translation \\ with Stage-wise Training}

\author{Fan Zhang\textsuperscript{\rm $1$,\rm $2$},
        Mei Tu\textsuperscript{\rm $2$},
        Sangha Kim\textsuperscript{\rm $2$},
        Song Liu\textsuperscript{\rm $2$},
        Jinyao Yan\textsuperscript{\rm $1$} \\
  \textsuperscript{\rm $1$} State Key Laboratory of Media Convergence and Communication, Communication University of China \\
  \textsuperscript{\rm $2$} Samsung Research China - Beijing (SRC-B) \\
  \texttt{\{zhang.fan, mei.tu, sangha01.kim, s0101.liu\}@samsung.com, jyan@cuc.edu.cn} \\}

\begin{document}
\maketitle
\begin{abstract}

Most multi-domain machine translation models rely on domain-annotated data.
Unfortunately, domain labels are usually unavailable in both training processes and real translation scenarios.
In this work, we propose a label-free multi-domain machine translation model which requires only a few or no domain-annotated data in training and no domain labels in inference.
Our model is composed of three parts: a backbone model, a domain discriminator taking responsibility to discriminate data from different domains, and a set of experts that transfer the decoded features from generic to specific.
We design a stage-wise training strategy and train the three parts sequentially.
To leverage the extra domain knowledge and improve the training stability, in the discriminator training stage, domain differences are modeled explicitly with clustering and distilled into the discriminator through a multi-classification task.
Meanwhile, the Gumbel-Max sampling is adopted as the routing scheme in the expert training stage to achieve the balance of each expert in specialization and generalization.
Experimental results on the German-to-English translation task show that our model significantly improves BLEU scores on six different domains and even outperforms most of the models trained with domain-annotated data.

\end{abstract}

\section{Introduction}

\begin{figure*}
\centering
\includegraphics[scale=0.20]{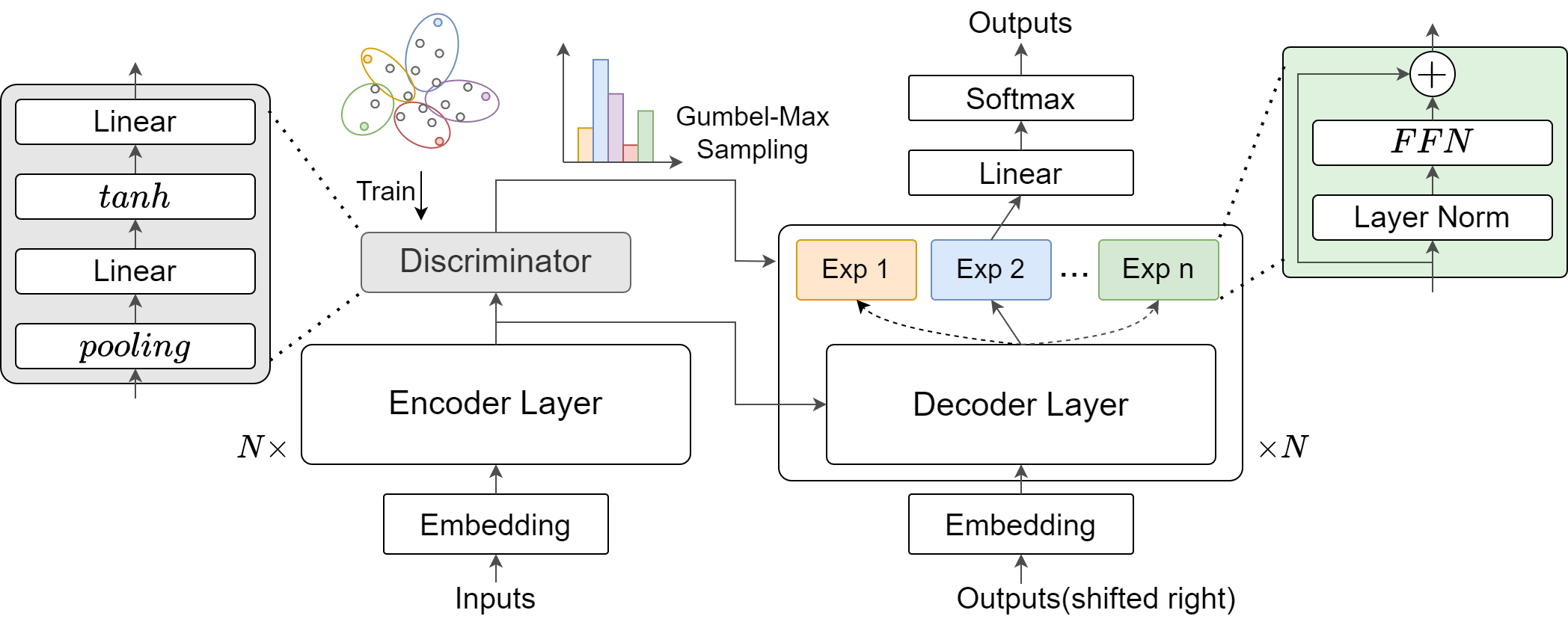}
\caption{Model architecture. Our model consists of an encoder-decoder-based backbone model, a discriminator, and a set of experts transplanted in the backbone model.
This model is trained sequentially with a stage-wise training strategy.
We model domain differences by clustering using as few domain-annotated data as possible and simulate the vague domain boundary with the routing scheme of the Gumbel-Max sampling.}
\label{fig:overview}
\end{figure*}

In recent years, neural machine translation (NMT), as one of the core tasks of neural language processing (NLP), has already been studied extensively with great progress \citep{vaswani2017attention, dabre2020survey}.
Texts of different domains usually have their own expression styles.
Domain diversity leads to heterogeneous data distribution of a domain-mixed dataset.
During the training of a generic NMT model, data from different domains tend to adjust model parameters to fitting their own distributions.
Such a global optimum sacrifices domain specialization, especially in those small domains.
Multi-domain machine translation (MDMT) still remains challenging \citep{pham2021revisiting}.

To help the model distinguish different domains, some works \citep{kobus2017domain, britz2017effective, zeng2018multi} tried to introduce domain knowledge during training.
Besides, some other works \citep{bapna2019simple, pham2020study} regarded MDMT as an extension of transfer learning in the machine translation area, and they tried to handle the domain shift problem by fine-tuning a generic model with in-domain datasets.
However, these two kinds of works require domain labels in training, and some even need to specify domains in inference.
Domain-label annotation is one of the biggest obstacles against these efforts.
On the one hand, domain labels are usually unavailable for large-scale bilingual parallel data.
On the other hand, the diversity of domain criteria makes the notion (boundary) of a domain vague, thus making the domain-label annotation difficult.
For instance, a news sentence describing medicine can belong to both the news domain and the medical domain.

Recently, previous works \citep{shazeer2017outrageously, lepikhin2020gshard, fedus2021switch, dai2022stablemoe} explored the mixture-of-experts (MoE) structure in NMT.
Since MoE implicitly learns to route data to suitable experts by the gating network, domain-label annotation is unnecessary when using MoE to handle MDMT.
These works have shown the powerful ability of MoE on dealing with diverse data distributions, but a training strategy should always be carefully designed to avoid training instability.
For instance, expert load imbalance may happen during training an MoE model: most data would have been routed to a small number of experts, meanwhile many other experts do not get sufficiently trained at all \citep{lepikhin2020gshard}.
Moreover, the routing fluctuation \citep{dai2022stablemoe} issue, i.e. the same data may be routed to different experts along with training, is also one of the factors leading to training instability.

In this work, we propose a label-free MDMT model which requires only a few or no domain-annotated data in training and no domain labels in inference.
Inspired by the concept of MoE, our model is composed of three parts: (i) a generic machine translation model as the backbone model; (ii) a domain discriminator (with respect to the gating network in the MoE structure); and (iii) a set of experts transplanted in the backbone model.
To guarantee training stability, we design a stage-wise training strategy and train the three parts individually in their respective stages.
When training the domain discriminator, domain differences are modeled explicitly by clustering using as few domain-annotated sentences as possible and distilled into the discriminator through a multi-classification task.
In other words, we (i) can leverage a little accessible extra domain knowledge easily and (ii) train the discriminator explicitly which enhances the training stability.
Meanwhile, considering that the domain boundary is vague, in the expert training stage, we adopt the Gumbel-Max sampling as the routing scheme and train the experts through the standard machine translation task without any domain-annotated data.
With the Gumbel-Max routing scheme, we achieve balance in the domain knowledge of each expert in specialization and generalization, which further improves the overall multi-domain translation quality.


Experimental results on the German-to-English translation task show that our model significantly improves BLEU scores on six different domains and even outperforms most of the models trained with domain-annotated data.
Further explorations of the domain discrimination ability show the success of the discriminator training manner.
Additionally, the ablation studies in the expert training stage demonstrate the effectiveness of the Gumbel-Max routing scheme.

\section{Related Works}

MDMT has been an object of research in the literature on neural language processing.
Some works \citep{kobus2017domain, britz2017effective, tars2018multi} show that injecting domain information either on the encoder side or on the decoder side can help the model improve domain translation quality.
Compared with the above works that introduce sentence-level domain information, some other works \citep{zeng2018multi, su2019exploring, jiang2020multi} inject token-level domain information to cope with multi-domain sentences.

Another strategy to handle MDMT is to regard it as an extension of transfer learning.
According to the parameter-efficient requirement, the lightweight adapter modules are well studied on various transfer learning tasks \citep{houlsby2019parameter, zhu2021counter}.
Some works \citep{bapna2019simple, pham2020study} transplant adapters into a generic model, and later fine-tune them with in-domain datasets.

However, all the above works need domain-annotated data, which means they are suffering from the data collection problem.
Actually, this topic has been widely studied for machine translation \citep{axelrod2011domain, van2017dynamic, aharoni2020unsupervised}, but it is not well integrated with the above works.

The MoE structure \citep{jacobs1991adaptive} has already been proposed for a long time.
In recent years, it has been widely studied in the machine translation area \citep{shazeer2017outrageously, shen2019mixture, lepikhin2020gshard, dai2022stablemoe}.
These works show that the training instability of the gating network requires a sophisticated training strategy.
Besides, high training cost has also become one of the reasons that prevent MoE from being widely studied.

\section{Model Architecture}

The core concept of MoE is using multiple experts to divide a problem space into homogeneous regions \citep{baldacchino2016variational}, which perfectly fits the MDMT task.
Following this clue, we design our model to be composed of three parts: (i) a generic machine translation model as the \textbf{backbone model}; (ii) a \textbf{discriminator} (with respect to the gating network in the MoE structure, which responsibility is to route data to suitable experts); and (iii) a set of \textbf{experts} transplanted in the decoder of the backbone model.
Figure \ref{fig:overview} illustrates the overall model architecture.

\textbf{The backbone model} is based on the encoder-decoder structure, where the encoder/decoder block is composed of a stack of several identical layers.
Given a source sentence $x=(x_1, ..., x_n)$, the encoder block maps it to a sequence of hidden states $h=(h_1, ..., h_n)$.
Then, $h$ is fed to the decoder block to generate an output sequence $y=(y_1, ..., y_m)$ with an auto-regressive process.

\textbf{The discriminator} makes use of the hidden states $h$ to discriminate different domain data by scoring predefined domain categories.
First, $h \in \mathbb{R}^{n \times d}$ is condensed to $\hat{h} \in \mathbb{R}^{d}$ by mean pooling on the sequence length dimension $n$,
\begin{equation}
    \hat{h} = Pooling(h)
\label{pooling}
\end{equation}
Then two linear transformations are introduced with a $tanh$ activation in between to compute category scores $s$,
\begin{equation}
    s = tanh(\hat{h} W_1 + b_1) W_2 + b_2
\label{ld}
\end{equation}
where $W_1 \in \mathbb{R}^{d \times d}$, $W_2 \in \mathbb{R}^{d \times K}$, $b_1 \in \mathbb{R}^{d}$ and $b_2 \in \mathbb{R}^{K}$ are the parameter matrices of the linear transformations and $K$ is the predefined category number.

\textbf{The experts} are transplanted in every decoder layer of the backbone model to transfer the decoded hidden states from generic to specific.
In each expert, the output $z_i$ of the $i$-th decoder layer is first normalized with layer normalization,
\begin{equation}
    \tilde{z_i} = LN(z_i)
\end{equation}
Then $\tilde{z_i}$ is fed to a fully connected feed-forward network $FFN(\cdot)$ \citep{vaswani2017attention}, followed by a residual connection, to obtain the expert output,
\begin{equation}
    o_i = FFN(\tilde{z_i}) + z_i
\end{equation}
In inference, the expert with the biggest category score in each decoder layer is activated.
Unlike inference, the Gumbel-Max sampling is adopted as the routing scheme in the expert training stage, which will be discussed in the next section.

\section{Stage-wise Training}

We design a stage-wise training strategy and train the three parts of our model sequentially.
It is worth noting that the parameters of all previous parts will be frozen when training the latter part.
The decoupled training process has brought us many benefits.
First, it is feasible to reuse a well-pretrained NMT model as the backbone model, which is a huge advantage in commercial usage since training an NMT model from scratch is relatively time-consuming.
Second, we can easily inject extra domain knowledge to help improve the performance of the discriminator, thus improving the translation quality.
Third, such a training process avoids the training instability problem caused by the MoE structure.
Next, we will discuss our training process in detail.

\subsection{Backbone model}

The backbone model is trained through a standard machine translation task.
Given a dataset of parallel text $D^{mt}=\{(x, y^*)\}^{N_t}_{i=1}$, the training objective is varying the trainable parameters $\theta$ to minimize the cross-entropy loss:
\begin{equation}
    \mathcal{L}_{mt}(\theta) = - \sum_{i=1}^{N_t}\sum_{t=1}^{m}log P(y^*_t|y^*_{1:t-1}, x; \theta)
\label{obj:mt}
\end{equation}
At this training stage, $\theta$ refers to the parameters of the backbone model.

\subsection{Discriminator}

\begin{figure}
\centering
\includegraphics[scale=0.3]{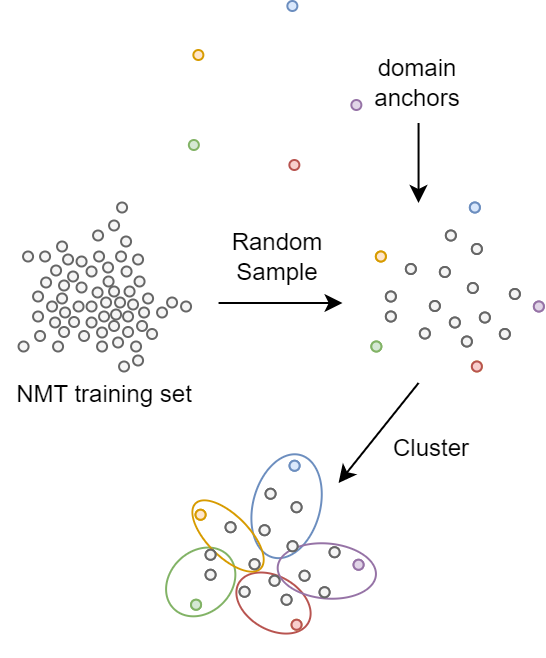}
\caption{Discriminator training set construction.}
\label{fig:clustering}
\end{figure}

Sentence features from the same domain are usually closer than those from different domains, so the domain differences can be modeled by clustering without any domain information \citep{aharoni2020unsupervised}.
Meanwhile, a small amount of in-domain sentences are readily accessible.
These sentences can be used for small domain augmentation when clustering.

In this training stage, a small sentence set is first prepared by sampling from $D^{mt}$ at random.
Optionally, a few in-domain sentences are injected into the sentence set as domain anchors if exist.
By the encoder of the backbone model, we convert these sentences into the condensed hidden states $\hat{h}$ (Eq. \ref{pooling}) as the sentence features and then cluster them into $K$ groups.
Each group is regarded as an independent domain category.
Figure \ref{fig:clustering} illustrates the above steps.
In the end, we distill this clustering result into the discriminator through a multi-classification task.

Let $D^d=\{(x, c)\}^{N_d}_{i=1}$ be the training set we construct above where $c$ is the one-hot vector of the domain category label, the goal in this training stage is minimizing the multi-classification loss:
\begin{equation}
    \mathcal{L}_{d}(\theta) = - \sum_{i=1}^{N_d}\sum_{j=1}^{K} c_j log (p_j)
\end{equation}
where
\begin{equation}
    p_j = \frac{e^{s_j}}{\sum_{k=1}^{K}e^{s_k}}
\end{equation}
and $\theta$ refers to the parameters in Eq. \ref{ld}.

\subsection{Experts}
\label{subsec:exp}

\begin{figure}
\centering
\includegraphics[scale=0.3]{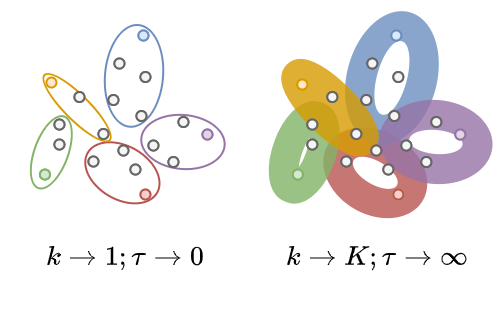}
\caption{Domain boundary simulation.
$k$ controls the boundary width while $\tau$ controls the domain probability distribution.
}
\label{fig:boundary}
\end{figure}

Since the domain boundary is vague, routing data to the expert with the highest score is reckless.
In the expert training stage, we propose routing sentences with the Gumbel-Max sampling scheme \citep{gumbel1954statistical, maddison2014sampling}.
While ensuring the specialization of domain knowledge, this routing scheme further improves the knowledge generalization of each expert.

Formally, given the category scores $s$, we focus on the categories with the top $k$ biggest scores and compute their relative probabilities,
\begin{equation}
    p = softmax(topk(s) / \tau)
\label{eq:gb}
\end{equation}
where temperature $\tau > 0$ is a hyper-parameter controlling the probability distribution.
The higher the $\tau$, the closer the probability distribution to the discrete uniform distribution.
On the contrary, it is closer to the one-hot distribution.
The activated expert is chosen as:
\begin{equation}
    e = arg\ max(G(p))
\end{equation}
where
\begin{equation}
    G(p) = log(p) + g
\end{equation}
and $g$ is a set of i.i.d samples that are drawn from Gumbel(0,1) distribution \citep{gumbel1954statistical}.
Please refer to Appendix \ref{subsec:appendix-gm} for detailed implementation.

The training objective in this stage is the same as the backbone model (Eq. \ref{obj:mt}), except that $\theta$ refers to the parameters of the experts.

\section{Experimental Settings}

To evaluate our model, we conduct a set of experiments on MDMT under the German-to-English translation task.
Different from most previous works \citep{kobus2017domain, pham2021revisiting} where the data scale of each domain is comparable, we mix several small domain datasets with a large generic dataset.
The translation quality is measured by the BLEU-4 \citep{papineni2002bleu} score.
In the following parts, we will describe our experimental settings in detail.

\subsection{Datasets}

We collect two datasets of German-English sentence pairs.
One is the standard WMT 2014 German-English dataset consisting of about $4.6$ million sentence pairs, which can be seen as a large generic domain (\textit{WMT}).
Another one is the multi-domain dataset from \citet{aharoni2020unsupervised} which is originally provided by \citet{koehn2017six}, including textual data in five diverse domains: IT-related text (\textit{IT}, manuals and localization files of open-source software), translations of the Koran (\textit{KOR}), legal text (\textit{LAW}, legislative text of the European Union), medical text (\textit{MED}, PDF documents from the European Medicines Agency), and subtitles (\textit{SUB}).
To explore model performance under the label-free scenario, we build a test set with random domain data, namely \textit{RND}, by sampling $500$ sentences from each domain test set at random with random domain labels.
In the data processing phase, we first use the Moses tokenizer \citep{koehn2007moses} for subword tokenization of both English and German sentences.
Then we learn and apply Byte-Pair Encoding \citep{sennrich2016neural} on all sentences jointly, while the number of the merge operation is set to $30,000$.

\begin{table}
\centering
\begin{tabular} {ccccc}
\hline
 & Train & Dev & Test \\
\hline
\textit{WMT} & $4,603,578$ & $2,001$ & $3,003$ \\
\textit{KOR} & $17,982$ & $2,000$ & $2,000$ \\
\textit{IT} & $222,927$ & $2,000$ & $2,000$ \\
\textit{MED} & $248,099$ & $2,000$ & $2,000$ \\
\textit{LAW} & $467,309$ & $2,000$ & $2,000$ \\
\textit{SUB} & $500,000$ & $2,000$ & $2,000$ \\
\textit{RND} & - & - & $3,000$ \\
\hline
\end{tabular}
\caption{Dataset statistics. \textit{RND} test set is built by choosing $500$ sentences from each domain test set at random with random domain labels.}
\label{tab:dataset}
\end{table}

\subsection{Implementations}

\begin{table*}
\centering
\begin{tabular}
{lrcccccccc}
\hline
 & $Param.$ & \textit{WMT} & \textit{KOR} & \textit{IT} & \textit{MED} & \textit{LAW} & \textit{SUB} & $AVG$ & \textit{RND} \\
\hline
\texttt{Mixed-nat} & $100M$ & 32.08 & 19.65 & 44.79 & 51.47 & 54.34 & 30.60 & 38.82 & 40.66 \\
\texttt{FT}$^{\star\diamond}$ & $[+500]M$ & 32.14 & \textbf{22.31} & \textbf{46.76} & \textbf{54.34} & \textbf{57.06} & \textbf{32.02} & \textbf{40.77} & 35.91 \\
\hline
\texttt{ADPT}$^{\star\diamond}$ & $[+9.5]M$ & 32.25 & \textbf{22.44} & 45.82 & 52.78 & 55.53 & \textbf{31.96} & 40.13 & 36.86 \\
\texttt{DC-tag}$^{\star\diamond}$ & $[+0]M$ & 31.98 & 20.76 & 45.08 & 52.10 & 54.86 & 31.35 & 39.36 & 38.22 \\
\texttt{DM}$^\star$ & $[+0]M$ & 31.65 & 18.81 & 43.69 & 50.85 & 53.22 & 30.68 & 38.15 & 40.25 \\
\texttt{ADM}$^\star$ & $[+0]M$ & 31.59 & 18.44 & 43.67 & 51.04 & 53.66 & 30.16 & 38.09 & 40.17 \\
\texttt{TTM}$^\star$ & $[+0]M$ & \textbf{32.26} & 19.11 & 44.34 & 51.63 & 54.23 & 31.54 & 38.85 & 40.57 \\
\texttt{SG-MoE} & $[+9.5]M$ & 31.83 & 21.04 & 45.45 & \textbf{53.48} & \textbf{56.79} & 31.30 & 39.98 & \textbf{41.74} \\
\texttt{Switch} & $[+9.5]M$ & 32.13 & 20.55 & 45.53 & 52.57 & 55.72 & 30.97 & 39.58 & 41.46 \\
\hline
\texttt{Ours-RS} & $[+9.7]M$ & \textbf{32.78} & 21.97 & \textbf{46.20} & 53.15 & 55.91 & 31.63 & \textbf{40.27} & \textbf{42.16} \\
\texttt{Ours-DI$^\star$} & $[+9.7]M$ & \textbf{32.45} & \textbf{22.27} & \textbf{46.29} & \textbf{53.54} & \textbf{56.10} & \textbf{31.72} & \textbf{40.40} & \textbf{42.22} \\
\hline
\end{tabular}
\caption{
Multi-domain translation performance.
The flags, $\star$ and $\diamond$, indicate the model needs domain information in training and inference, respectively.
The average BLEU scores of the six domains (\textit{WMT}, \textit{KOR}, \textit{IT}, \textit{MED}, \textit{LAW}, \textit{SUB}) are listed in $AVG$ column.
\textit{RND} reflects the model performance without domain labels (or with random domain labels in the label-required models).
The top three BLEU scores in each column are shown in bold.
Please note that compared with other models with flag $\star$, \texttt{Ours-DI} requires very few domain-annotate data.
}
\label{tab:main}
\end{table*}

We use the standard Transformer \citep{vaswani2017attention} implemented in Fairseq \citep{ott2019fairseq} with $6$ layers, $512$ embedding dimensions, and $8$ attention heads as the base model structure.
All comparison models are implemented with the base model structure.
Next, we will introduce these models briefly.

\textbf{Baseline models}.
We first train a standard NMT model using the naturally mixed dataset, referred to as \texttt{Mixed-nat}, which is also our backbone model.
Furthermore, we then fine-tune \texttt{Mixed-nat} for every domain by their in-domain datasets.
The six fine-tuned models are considered as an ensemble, namely \texttt{FT}.

\textbf{Adapters} (\texttt{ADPT}) \citep{bapna2019simple}.
The experts (adapters) are injected in layers of both encoder and decoder of the pre-trained \texttt{Mixed-nat}.
Then, for each domain, its experts are fine-tuned with the in-domain dataset independently.

\textbf{Domain Control with Tag} (\texttt{DC-Tag}) \citep{kobus2017domain}.
\texttt{DC-Tag} introduces domain information as an additional token for each source sentence in both training and inference.

\textbf{Discriminative Mixing} (\texttt{DM}), \textbf{Adversarial Discriminative Mixing} (\texttt{ADM}) and \textbf{Target Token Mixing} (\texttt{TTM}).
The three methods come from the same work \citep{britz2017effective}.
\texttt{DM} and \texttt{ADM} minimize the joint training objective of domain classification task and translation task by using different learning strategies, while \texttt{TTM} adds domain tags in the front of target sentences to make the model predict domain tag as a regular token.

\textbf{Sparsely-gated mixture-of-experts} (\texttt{SG-MoE}) \citep{shazeer2017outrageously} and \textbf{Switch transformer} (\texttt{Switch}) \citep{fedus2021switch}.
\texttt{SG-MoE} introduces sparsely gated MoE layers with the noisy top-k gating scheme, which activates $k>1$ experts per time to obtain nonzero derivatives in back-propagation.
It also introduces auxiliary losses to deal with the expert load imbalance.
In practice, we set $k=2$ for \texttt{SG-MoE}.
\texttt{Switch} is another MoE method that activates only one expert per time.
To avoid the expert load imbalance when training the gating network, it introduces a capacity factor and an auxiliary load balancing loss.
In our implementations\footnote{The losses in our implementations are referred to that in https://github.com/laekov/fastmoe .}, different from the original works that use token-level expert routing schemes, we adopt sentence-level routing scheme and only introduce experts in the decoder layers to consist with ours.

\textbf{Ours}.
In the discriminator training stage, we sample $50,000$ sentences from the mixed NMT training set at random and use sentences of the Dev sets as the domain anchors.
In contrast, we also train the discriminator with only the random sampled (RS) sentences to show the influence of the injected domain information (DI).
The models that the discriminator trained without and with the domain anchors are referred to as \texttt{Ours-RS} and \texttt{Ours-DI}, respectively.
We choose Gaussian Mixture Model (GMM) as our clustering method.
The expert number $K$ is set to $12$ for all MoE models (\texttt{SG-MoE}, \texttt{Switch} and Ours), while the inner dimension of the experts is set to $128$ for all MoE models and \texttt{ADPT}.
In the expert training stage, we set the hyper-parameters of the Gumbel-Max routing scheme $k=4$ and $\tau = 1.0$.
The Gumbel-Max routing scheme is shut down in inference with $k=1$.



\section{Results and Discussion}

\subsection{Multi-domain machine translation}

Most of the MDMT models rely on domain-annotated data in training, and some even require domain labels in inference.
We test the translation performance of the MDMT models on the six domains (\textit{WMT}, \textit{KOR}, \textit{IT}, \textit{MED}, \textit{LAW}, \textit{SUB}) and report their BLEU scores in Table \ref{tab:main}.
BLEU scores on the \textit{RND} test set reflect the translation performance under the label-free scenario.
To show the differences in model size, we also present the number of parameters ($Param.$) in Table \ref{tab:main}.

As shown in Table \ref{tab:main}, our models (\texttt{Ours-RS} and \texttt{Ours-DI}) outperform all other MDMT models on overall multi-domain translation measurements ($AVG$ and \textit{RND}).
Although \texttt{FT} shows a better average BLEU score, it requires an extremely large model size in the meantime.
In inference, label-required models (i.e. models with flag $\diamond$) perform well with domain labels (see $AVG$).
However, when they are confronted with the scenario that domain labels are unavailable (or probably not correct), their performance is significantly poorer than others (see \textit{RND}).
In contrast, the outstanding performance of the MoE models (i.e. \texttt{SG-MoE}, \texttt{Switch}, \texttt{Ours-RS} and \texttt{Ours-DI}) on the \textit{RND} test set illustrates the powerful ability of MoE on dealing with the MDMT problem.
Given domain labels, \texttt{ADPT} can be regarded as an MoE model with oracle discrimination ability.
But its expert number is limited by the known domain number, and the small domains will not take benefits from the big generic dataset (i.e. the \textit{WMT} training set in our experiments).
Results show that although it has oracle discrimination ability, its model performance on some small domains is poorer than ours.

Comparing \texttt{Ours-DI} with \texttt{Ours-RS}, we find that \texttt{Ours-DI} shows steady improvement on those small domains.
The improvement is attributed to the improved domain discrimination ability of the discriminator, which will be discussed in the following part.
It also indicates that the ability to flexibly inject domain information is necessary for an MDMT model.

To evaluate the dataset robustness of our model, we further train our model with only the \textit{WMT} training set and on the English-to-German translation direction, respectively.
The experimental results are reported in Appendix \ref{subsec:appendix-dataset}.

\subsection{Domain discrimination ability}

\begin{figure}[htb]
\centering
\includegraphics[scale=0.7]{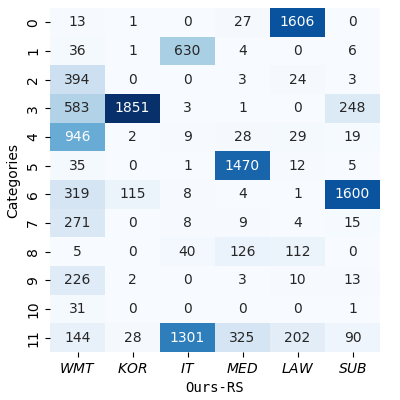}
\caption{Routing statistics of \texttt{Ours-RS} on the test sets.}
\label{fig:rs}
\end{figure}

\begin{figure}[htb]
\centering
\includegraphics[scale=0.7]{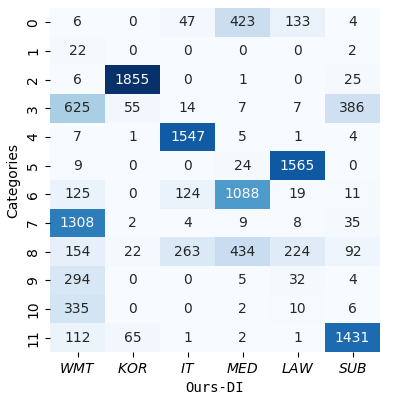}
\caption{Routing statistics of \texttt{Ours-DI} on the test sets.}
\label{fig:di}
\end{figure}

To analyze the impact of the clustering results on domain discrimination ability and ultimately on the translation performance, we conduct a set of experiments in the discriminator training stage.

First, we count the routing results of \texttt{Ours-RS} and \texttt{Ours-DI} on the test sets (Figure \ref{fig:rs} and Figure \ref{fig:di}).
Based on the statistics, we roughly measure the domain discrimination ability by two metrics.
One is the category purity score $PUR$,
\begin{equation}
    PUR = \frac{1}{U} \sum_{i=1}^{K} u_i^{max}
\end{equation}
where $U$ is the total number of the test cases, $K$ is the number of the categories (NOT the number of the test domains), and $u_i^{max}$ is the maximum number of $i$-th category.
The other one is the normalized mutual information $NMI$ \citep{danon2005comparing} score between true domain labels and the predicted category labels, as implemented in scikit-learn \citep{pedregosa2011scikit}.
The two metrics measure the mixing degree of different domains in a category.
The higher the $PUR$ and the $NMI$, the better the domain discrimination ability.

Results in Table \ref{tab:disc} show that the domain discrimination ability of our discriminator is improved after injecting a few domain-annotated data.
When we investigate the routing statistical results of \texttt{Ours-RS} in Figure \ref{fig:rs}, we find that although most sentences in the same domain are routed to the same category, the category purity is relatively low.
Take \textit{KOR} as an example, most of its sentences are routed to the $3$-rd category, but this category also contains many sentences from \textit{WMT} and \textit{SUB}.
In contrast, routing statistics of \texttt{Ours-DI} show higher purity of clusters related to small domains.

\begin{table}
\centering
\begin{tabular} {ccc}
\hline
 & $PUR$ & $NMI$ \\
\hline
\texttt{Ours-RS} & 0.8038 & 0.5976 \\
\texttt{Ours-DI} & \textbf{0.8403} & \textbf{0.6455} \\
\hline
\end{tabular}
\caption{Measurements of the domain discrimination ability.}
\label{tab:disc}
\end{table}

\begin{table}
\centering
\begin{tabular} {lccc}
\hline
 & GMM & K-Means & Birch \\
\hline
\textit{WMT} & 32.45 & \textbf{32.93} & 32.77 \\
\textit{KOR} & 22.27 & 22.09 & \textbf{22.35} \\
\textit{IT} & \textbf{46.29} & 46.23 & 46.08 \\
\textit{MED} & 53.54 & \textbf{53.78} & 53.58 \\
\textit{LAW} & 56.10 & 56.01 & \textbf{56.23} \\
\textit{SUB} & \textbf{31.72} & 31.45 & 31.17 \\
$AVG$ & 40.40 & \textbf{40.42} & 40.36 \\
\textit{RND} & \textbf{42.22} & 42.12 & 42.00 \\
\hline
\end{tabular}
\caption{BLEU scores of different clustering methods.}
\label{tab:cluster}
\end{table}

We also try another two clustering methods (K-Means and Birch) to model the domain differences in the discriminator training stage.
The final BLEU scores based on the three clustering methods are listed in Table \ref{tab:cluster}.
Results show that the three clustering methods have their own advantages and disadvantages for different domains.
It is difficult to define which one is better.

Another interesting observation is that after distilling the clustering result into the discriminator, the domain discrimination ability of the discriminator is even better than that of the clustering model.
Measurements of the domain discrimination abilities are reported in Table \ref{tab:disc-ability}.
We find that there are steady improvements on both $PUR$ and $NMI$ for all three clustering methods.
Previous works \citep{tian2017deepcluster, caron2018deep, aharoni2020unsupervised} also have reported similar observations on different tasks.

\begin{table}
\centering
\begin{tabular} {llcc}
\hline
 & & $PUR$ & $NMI$ \\
\hline
\multirow{2}{*}{GMM} & cluster & 0.8365 & 0.6377 \\
 & distill & \textbf{0.8403} & \textbf{0.6402} \\
\hline
\multirow{2}{*}{K-Means} & cluster & 0.8483 & 0.6216 \\
 & distill & \textbf{0.8508} & \textbf{0.6242} \\
\hline
\multirow{2}{*}{Birch} & cluster & 0.8068 & 0.5967 \\
 & distill & \textbf{0.8133} & \textbf{0.6087} \\
\hline
\end{tabular}
\caption{Clustering models vs discriminators. `cluster' means routing by clustering models, while `distill' means routing by discriminators (trained by distilling knowledge from clustering models)}
\label{tab:disc-ability}
\end{table}

\begin{figure}
\centering
\includegraphics[scale=0.5]{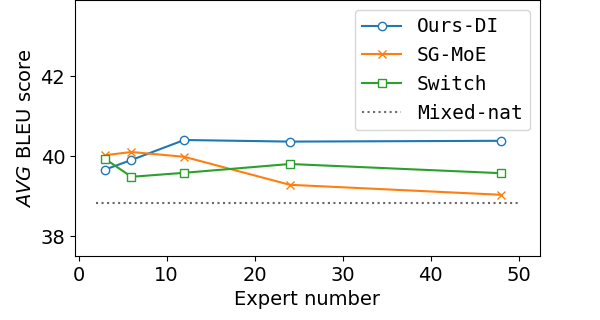}
\caption{Expert number analysis. We set the expert number $K$ to $3$, $6$, $12$, $24$, and $48$ on each MoE method to explore its impact.}
\label{fig:expnum}
\end{figure}

\subsection{Number of experts}

Based on \texttt{Ours-DI}, we vary the expert number $K$ (i.e. the domain category number) and explore its impact.
Meanwhile, we also explore this impact on the other two MoE methods (\texttt{SG-MoE} and \texttt{Switch}).
Figure \ref{fig:expnum} shows our method can improve the performance of the backbone model (\texttt{Mixed-nat}) even when $K$ is smaller than the known domain number.
With the increase of $K$, too many experts will not bring significant improvement under our experimental settings.
Although \texttt{SG-MoE} and \texttt{Switch} introduce auxiliary losses to alleviate expert load imbalance, the training instability problem still exists.
When the number of experts increases, their $AVG$ BLEU scores decrease instead.

\subsection{Routing scheme}

In the expert training stage, we adopt the Gumbel-Max routing scheme to simulate the vague domain boundary.
The two hyper-parameters, the category candidate number $k$ and the temperature $\tau$, control the mixing degree between different categories (w.r.t. Eq. \ref{eq:gb}).
We experiment with adjusting them in the expert training stage.
Experimental results are reported in Table \ref{tab:gm-t} and Table \ref{tab:gm-k}.

\begin{table}
\centering
\begin{tabular} {lcc}
\hline
 & $AVG$ & \textit{RND} \\
\hline
\texttt{Mixed-nat} & 38.82 & 40.66 \\
\hline
$\tau \rightarrow 0.0$ & 40.10 & 41.96 \\
$\tau = 0.1$ & 40.31 & 42.12 \\
$\tau = 1.0$ & \textbf{40.40} & \textbf{42.22} \\
$\tau = 10.0$ & 40.11 & 42.17 \\
\hline
\end{tabular}
\caption{Hyper-parameter $\tau$ analysis. $\tau \rightarrow 0.0$ is equivalent to shutting down the Gumbel-Max routing scheme.}
\label{tab:gm-t}
\end{table}

\begin{table}
\centering
\begin{tabular} {lcc}
\hline
 & $AVG$ & \textit{RND} \\
\hline
\texttt{Mixed-nat} & 38.82 & 40.66 \\
\hline
$k = 1$ & 40.10 & 41.96 \\
$k = 2$ & 40.39 & 42.11 \\
$k = 4$ & \textbf{40.40} & \textbf{42.22} \\
$k = 8$ & 40.31 & 42.06 \\
$k = 12$ & 40.17 & 41.98 \\
\hline
\end{tabular}
\caption{Hyper-parameter $k$ analysis. $k = 1$ is equivalent to shutting down the Gumbel-Max routing scheme.}
\label{tab:gm-k}
\end{table}

In Table \ref{tab:gm-t}, we fix $k$ to $4$ and vary $\tau$ to analyze the difference.
Meanwhile, the experimental settings in Table \ref{tab:gm-k} are that $\tau$ is fixed to $1.0$ and $k$ is varied.
Both $\tau \rightarrow 0.0$ and $k = 1$ are equivalent to shutting down the Gumbel-Max routing scheme.
Results show that the moderate values $\tau = 1.0$ and $k = 4$ perform better than the other settings.
It demonstrates that there is a balance in the domain knowledge of each expert in specialization and generalization.

\section{Conclusion}

This paper proposes a label-free MDMT model, which requires only a few or no domain-annotated data in training and no domain labels in inference.
Inspired by the concept of MoE, the model is composed of three parts and these parts are trained sequentially via a stage-wise training strategy.
Domain differences are modeled explicitly with clustering and distilled into the discriminator through a multi-classification task.
In addition, the Gumbel-Max sampling is adopted as the routing scheme in the expert training stage to achieve a balance of each expert in specialization and generalization.
Experimental results show the model significantly improves BLEU scores on six different domains and even outperforms most of the comparison models trained with domain-annotated data.

\section*{Limitations}

In this work, in order to tackle the problem of domain-label annotation, we use a clustering method to model domain differences with as few domain-annotated sentences as possible.
Empirical results show that although our performance is one of the best among the comparison models, there is still a small gap between ours and the fully fine-tuned model (\texttt{FT}).
To further improve our performance, it might be feasible to train a more powerful discriminator with more domain-annotated data or to use experts that have larger capacities, but that goes against our original intention.
Exploring a better approach than clustering to model domain differences by few to no domain-annotated sentences should be an interesting direction.
Additionally, the Gumbel-Max sampling scheme helps us improve the model performance, but its two hyper-parameters are fixed empirically in the current version.
In future work, adjusting these two hyper-parameters automatically according to the number of experts (categories) and the characteristics of the training set may be better.


\bibliography{anthology,custom}
\bibliographystyle{acl_natbib}

\appendix

\section{Appendices}
\label{sec:appendix}

\begin{algorithm*}[htb]
\caption{Gumbel-Max sampling of PyTorch version}
\SetKwData{Left}{left}
\SetKwData{This}{this}
\SetKwData{Up}{up}
\SetKwFunction{Union}{Union}
\SetKwFunction{FindCompress}{FindCompress}
\SetKwInOut{Parameter}{params}
\Parameter{Category scores $S$; Category candidate number $k$; Temperature $\tau$; Activated expert indices $E$.}
\BlankLine
\textbf{import} $torch$; \\
\textbf{import} $torch.nn.functional$ as $F$; \\
\If{$k<=1$}
{$E = torch.argmax(S)$; \\
\textbf{return} $E$; }
$topk\_val, topk\_idx = torch.topk(S, k=k, dim=1)$;\\
$topk\_val \ /= \tau$; \\
$log\_probs = torch.log(F.softmax(topk\_val, dim=1))$; \\
$g = F.gumbel\_softmax(log\_probs, dim=1)$; \\
$sampled = torch.argmax(g, dim=1, keepdim=True)$; \\
$E = torch.gather(topk\_idx, 1, sampled).squeeze()$; \\
\textbf{return} $E$;
\label{algo:gm}
\end{algorithm*}

\subsection{Training details}

In any training phase, we use the Adam optimizer \citep{kingma2014adam} with $\beta _1=0.9$, $\beta _2=0.98$ and $\epsilon = 10^{-9}$.
For translation optimization, we use the Noam decay as the learning rate scheduler with $4000$ warmup steps and a learning rate of $0.0007$.
With a batch size of $8192$ in the token level and the update frequency of $10$ on $2$ V100 GPUs, the maximum update number of training is set to $100k$, while that of fine-tuning is set to $10k$ with the early stopping strategy.
The maximum update number of the discriminator training stage is set to $6k$.
In inference, the beam size is set to $5$ for all models.

\subsection{Clustering details}
\label{subsec:appendix-cluster}

We choose Gaussian Mixture Model (GMM), K-Means and Birch implemented in scikit-learn \citep{pedregosa2011scikit} as our clustering approaches.
The convariance type of GMM is set to `full', while all other settings are set by default.
Before clustering, we perform dimensionality reduction with Principal Components Analysis (PCA) to reduce the vector dimension of the sentence representations from $512$ to $64$.
When testing the clustering performance on the test set, we reuse both the clustering model and the dimensionality reduction model generated by the training set.

\subsection{Gumbel-Max sampling}
\label{subsec:appendix-gm}

We implement the Gumbel-Max sampling strategy with PyTorch \citep{paszke2019pytorch} of version 1.10.1+cu102.
Implementation details are shown in Algorithm \ref{algo:gm}.
It is worth noting that the category scores $S$ and the activated expert indices $E$ are at batch-level compared with that in subsection \ref{subsec:exp}.

\subsection{Dataset robustness}
\label{subsec:appendix-dataset}

We conduct a set of experiments over different datasets to show the dataset robustness of our model.
First, we use only the \textit{WMT} training set as the translation training set and use only the random sampled sentences in the discriminator training stage (\texttt{Ours-RS}).
In this scenario, there is no concept of the domain at all.
Results in Table \ref{tab:only-wmt} show that without any in-domain data, \texttt{Ours-RS} also shows steady improvement compared with the backbone model.

\begin{table}
\centering
\begin{tabular} {lcc}
\hline
 & \texttt{Backbone} & \texttt{Ours-RS} \\
\hline
\textit{WMT} & 32.29 & \textbf{32.68} \\
\textit{KOR} & 13.24 & \textbf{13.41} \\
\textit{IT} & 35.91 & \textbf{36.04} \\
\textit{MED} & \textbf{38.29} & 37.87 \\
\textit{LAW} & 41.60 & \textbf{42.02} \\
\textit{SUB} & 23.76 & \textbf{24.07} \\
$AVG$ & 30.85 & \textbf{31.02} \\
\textit{RND} & 31.92 & \textbf{32.13} \\
\hline
\end{tabular}
\caption{BLEU scores with only \textit{WMT} training set.}
\label{tab:only-wmt}
\end{table}

Then, we test our model on the English-to-German translation direction.
Despite the translation direction, the model implementations and the training settings are all the same with the German-to-English translation task.
Results in Table \ref{tab:e2d} further demonstrate that the conclusions about our model performance in the main part can also be concluded on the English-to-German translation direction.

\begin{table}
\centering
\begin{tabular} {lccc}
\hline
 & \texttt{Mixed-nat} & \texttt{Ours-RS} & \texttt{Ours-DI} \\
\hline
\textit{WMT} & 27.06 & 27.23 & \textbf{27.42} \\
\textit{KOR} & 19.17 & 21.05 & \textbf{22.25} \\
\textit{IT} & 38.63 & 39.85 & \textbf{39.94} \\
\textit{MED} & 48.15 & 48.92 & \textbf{49.25} \\
\textit{LAW} & 48.97 & 50.10 & \textbf{50.39} \\
\textit{SUB} & 25.38 & 25.67 & \textbf{25.93} \\
$AVG$ & 34.56 & 35.47 & \textbf{35.86} \\
\textit{RND} & 36.52 & 37.54 & \textbf{37.93} \\
\hline
\end{tabular}
\caption{BLEU scores on the English-to-German translation direction.}
\label{tab:e2d}
\end{table}

\subsection{Routing statistics}

\begin{figure}[p]
\centering
\includegraphics[scale=0.7]{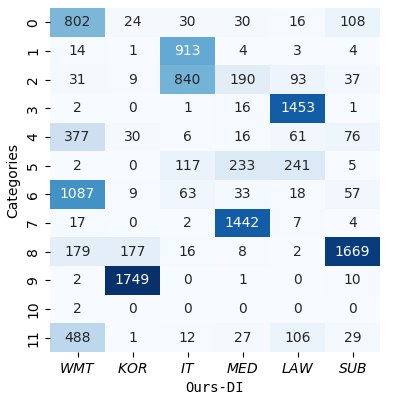}
\caption{Routing statistics of \texttt{Ours-DI} with K-Means clustering method on the test sets.}
\label{fig:di-km}
\end{figure}

\begin{figure}[p]
\centering
\includegraphics[scale=0.7]{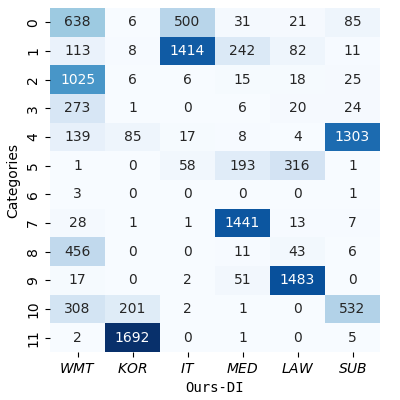}
\caption{Routing statistics of \texttt{Ours-DI} with Birch clustering method on the test sets.}
\label{fig:di-birch}
\end{figure}

We report the routing statistics of \texttt{Ours-DI} on the test sets with K-Means and Birch clustering methods in Figure \ref{fig:di-km} and Figure \ref{fig:di-birch}, respectively.

\subsection{Detailed BLEU scores}
We report the detailed BLEU scores of the expert number analysis and the hyper-parameter search experiments in Table \ref{tab:exp-num-detail} and Table \ref{tab:tau-k-detail}, respectively.

\begin{table*}
\centering
\begin{tabular}
{llccccccccc}
\hline
 & & \textit{WMT} & \textit{KOR} & \textit{IT} & \textit{MED} & \textit{LAW} & \textit{SUB} & $AVG$ & \textit{RND} \\
\hline
\multicolumn{2}{l}{\texttt{Mixed-nat}} & 32.08 & 19.65 & 44.79 & 51.47 & 54.34 & 30.60 & 38.82 & 40.66 \\
\hline
\multirow{5}{*}{\texttt{Ours-DI}} & $K=3$ & \textbf{32.58} & 20.58 & 45.47 & 52.44 & 55.48 & 31.41 & 39.66 & 41.56 \\
 & $K=6$ & 32.48 & 20.86 & 45.93 & 52.78 & 55.97 & 31.38 & 39.90 & 41.73 \\
 & $K=12$ & 32.45 & 22.27 & 46.29 & 53.54 & 56.10 & \textbf{31.72} & \textbf{40.40} & 42.22 \\
 & $K=24$ & 32.33 & \textbf{22.36} & 46.39 & 53.40 & 56.22 & 31.47 & 40.36 & \textbf{42.37} \\
 & $K=48$ & 32.01 & 21.73 & \textbf{46.63} & \textbf{54.39} & \textbf{56.68} & 30.86 & 40.38 & 42.23 \\
\hline
\multirow{5}{*}{\texttt{SG-MoE}} & $K=3$ & 32.19 & 20.86 & 45.30 & 53.47 & 56.31 & \textbf{31.96} & 40.02 & 41.83 \\
 & $K=6$ & \textbf{32.33} & 21.02 & \textbf{45.45} & \textbf{53.59} & 56.77 & 31.44 & \textbf{40.10} & \textbf{41.97} \\
 & $K=12$ & 31.83 & \textbf{21.04} & \textbf{45.45} & 53.48 & \textbf{56.79} & 31.30 & 39.98 & 41.74 \\
 & $K=24$ & 31.57 & 20.53 & 44.66 & 51.91 & 55.56 & 31.43 & 39.28 & 41.38 \\
 & $K=48$ & 31.53 & 20.33 & 44.51 & 52.05 & 55.41 & 30.34 & 39.03 & 40.76 \\
\hline
\multirow{5}{*}{\texttt{Switch}} & $K=3$ & 31.95 & \textbf{21.02} & \textbf{45.95} & 52.72 & \textbf{56.53} & \textbf{31.40} & \textbf{39.93} & \textbf{42.00} \\
 & $K=6$ & 32.12 & 20.66 & 45.29 & 52.18 & 55.98 & 30.63 & 39.48 & 41.17 \\
 & $K=12$ & \textbf{32.13} & 20.55 & 45.53 & 52.57 & 55.72 & 30.97 & 39.58 & 41.46 \\
 & $K=24$ & 31.85 & 20.59 & 45.54 & \textbf{53.40} & 56.40 & 31.04 & 39.80 & 41.81 \\
 & $K=48$ & 31.49 & 20.84 & 45.64 & 52.09 & 56.27 & 31.10 & 39.57 & 41.49 \\
\hline
\end{tabular}
\caption{Detailed BLEU scores of the expert number analysis. The highest BLEU score of each test set in each method is shown in bold.}
\label{tab:exp-num-detail}
\end{table*}

\begin{table*}
\centering
\begin{tabular}
{llccccccccc}
\hline
 & & \textit{WMT} & \textit{KOR} & \textit{IT} & \textit{MED} & \textit{LAW} & \textit{SUB} & $AVG$ & \textit{RND} \\
\hline
\multicolumn{2}{l}{\texttt{Mixed-nat}} & 32.08 & 19.65 & 44.79 & 51.47 & 54.34 & 30.60 & 38.82 & 40.66 \\
\hline
\multirow{4}{*}{\texttt{Ours-DI}} & $\tau \rightarrow 0.0$ & 32.35 & 22.05 & 46.18 & 52.78 & 55.84 & 31.42 & 40.10 & 41.96 \\
 & $\tau=0.1$ & \textbf{32.63} & 22.11 & 46.30 & 53.01 & \textbf{56.25} & 31.54 & 40.31 & 42.12 \\
 & $\tau=1.0$ & 32.45 & \textbf{22.27} & 46.29 & \textbf{53.54} & 56.10 & \textbf{31.72} & \textbf{40.40} & \textbf{42.22} \\
 & $\tau=10.0$ & 32.58 & 21.97 & \textbf{46.41} & 52.56 & 55.67 & 31.46 & 40.11 & 42.17 \\
\hline
\multirow{5}{*}{\texttt{Ours-DI}} & $k=1$ & 32.35 & 22.05 & 46.18 & 52.78 & 55.84 & 31.42 & 40.10 & 41.96 \\
 & $k=2$ & 32.63 & \textbf{22.68} & \textbf{46.33} & 53.50 & 55.93 & 31.24 & 40.38 & 42.11 \\
 & $k=4$ & 32.45 & 22.27 & 46.29 & \textbf{53.54} & \textbf{56.10} & \textbf{31.72} & \textbf{40.40} & \textbf{42.22} \\
 & $k=8$ & 32.58 & 22.52 & 46.09 & 53.12 & 56.01 & 31.55 & 40.31 & 42.06 \\
 & $k=12$ & \textbf{32.68} & 21.92 & 46.14 & 53.34 & 55.71 & 31.22 & 40.17 & 41.98 \\
\hline
\end{tabular}
\caption{Detailed BLEU scores of the hyper-parameter search experiments. The highest BLEU score of each test set in each hyper-parameter search is shown in bold.}
\label{tab:tau-k-detail}
\end{table*}

\end{document}